\documentclass[10pt,twocolumn,letterpaper]{article}

\usepackage{iccv}
\usepackage{times}
\usepackage{epsfig}
\usepackage{graphicx}
\usepackage{amsmath}
\usepackage{amssymb}
\DeclareMathOperator*{\argmin } {min}

\usepackage{epstopdf}
\usepackage{amssymb}
\usepackage{verbatim}
\usepackage{placeins}
\usepackage{bbm}
\usepackage{algorithm}
\usepackage{algpseudocode}
\usepackage{subfig}
\usepackage{tabularx}
\usepackage{array}
\newcolumntype{M}[1]{>{\centering\arraybackslash}m{#1}}
\graphicspath{{./images/}}
\usepackage{multirow}
\usepackage{mathtools}
\usepackage{enumitem}

\usepackage[pagebackref=true,breaklinks=true,letterpaper=true,colorlinks,bookmarks=false]{hyperref}

 \iccvfinalcopy % *** Uncomment this line for the final submission

\def\iccvPaperID{3220} % *** Enter the ICCV Paper ID here

\ificcvfinal\fi
\begin{document}
\setlength{\belowdisplayskip}{3.5pt} 
\setlength{\belowdisplayshortskip}{3.5pt}
\setlength{\abovedisplayskip}{3.5pt} 
\setlength{\abovedisplayshortskip}{3.5pt}

%%%%%%%%% TITLE
\title{Face Sketch Matching via Coupled Deep Transform Learning}

\author{Shruti Nagpal$^1$, Maneet Singh$^1$, Richa Singh$^{1,2}$, Mayank Vatsa$^{1,2}$, Afzel Noore$^2$, and Angshul Majumdar$^1$\\
$^1$IIIT-Delhi, India, $ ^2$West Virginia University\\
\{\tt\small shrutin, maneets, rsingh, mayank, angshul\}@iiitd.ac.in, afzel.noore@mail.wvu.edu
}

\maketitle
%\thispagestyle{empty}

%%%%%%%%% ABSTRACT
\begin{abstract}

Face sketch to digital image matching is an important challenge of face recognition that involves matching across different domains. Current research efforts have primarily focused on extracting domain invariant representations or learning a mapping from one domain to the other. In this research, we propose a novel transform learning based approach termed as DeepTransformer, which learns a transformation and mapping function between the features of two domains. The proposed formulation is independent of the input information and can be applied with any existing learned or hand-crafted feature. Since the mapping function is directional in nature, we propose two variants of DeepTransformer: (i) semi-coupled and (ii) symmetrically-coupled deep transform learning. This research also uses a novel IIIT-D Composite Sketch with Age (CSA) variations database which contains sketch images of 150 subjects along with age-separated digital photos. The performance of the proposed models is evaluated on a novel application of sketch-to-sketch matching, along with sketch-to-digital photo matching. Experimental results demonstrate the robustness of the proposed models in comparison to existing state-of-the-art sketch matching algorithms and a commercial face recognition system.
\end{abstract}

%%%%%%%%% BODY TEXT
\section{Introduction}
\begin{figure}
\centering
\subfloat[Viewed sketches]{\includegraphics[height=1.5in]{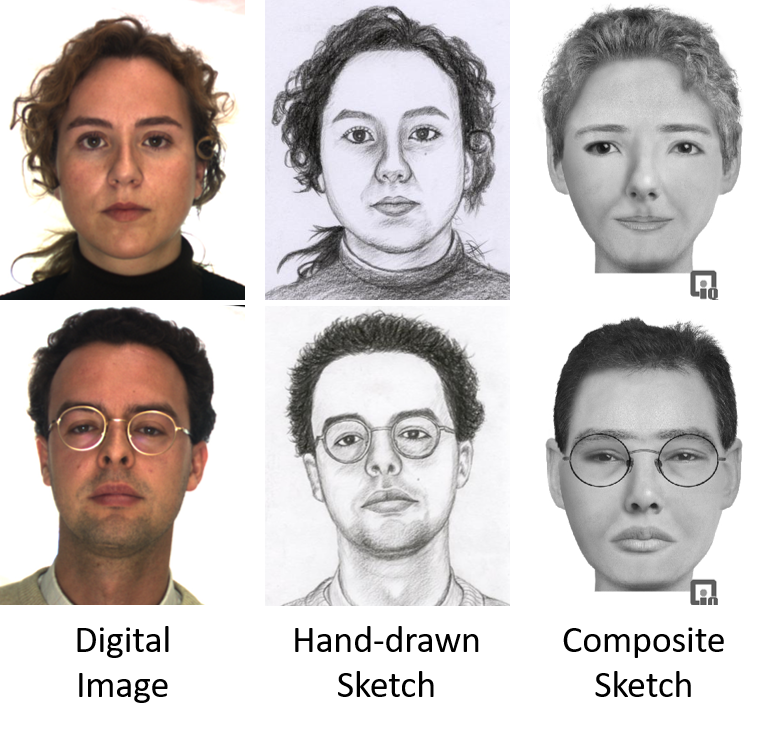}}
\hspace{15pt}\subfloat[Forensic sketches]{\includegraphics[height=1.5in]{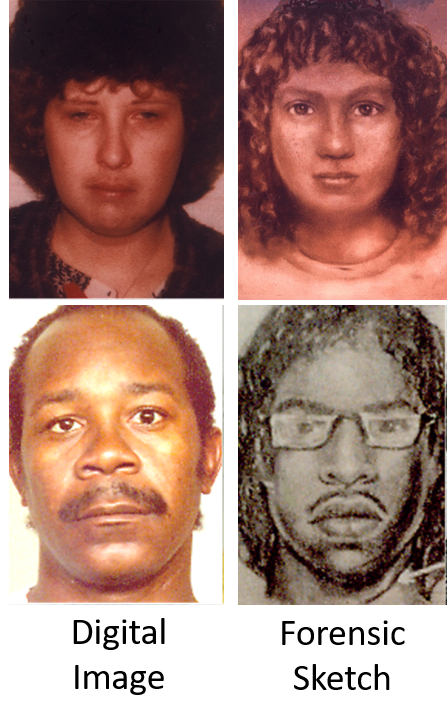}}
\caption{Illustrating the variations in the information content of digital images and different types of sketches.}
\label{intro1}
\vspace{-15pt}
\end{figure}

\begin{table*}[t]
\begin{center}
 \small
\begin{tabular}{|M{1.4cm}|m{3.2cm}|m{5.23 cm}|m{6 cm}|}
\hline
\textbf{Sketch} &\textbf{Authors (Year)} & \textbf{Feature Extraction} & \textbf{Classification} \\
\hline\hline
\multirow{7}{*}{Hand-drawn} & Bhatt \textit{et al.} \cite{iiitdSketch} (2012) & Proposed MCWLD & Memetically optimized chi-squared distance \\
\cline{2-4}
& Khan \textit{et al.} \cite{dicta} (2012) & Facial Self Similarity descriptor & Nearest neighbor classifier\\
\cline{2-4}
& Mignon \textit{et al.}\cite{mignon2012} (2012) & \multicolumn{2}{c|} {Proposed Cross modal metric learning (CMML)} \\
\cline{2-4}
& Klare \textit{et al.} \cite{jainHFR} (2013) & MLBP, SIFT + Heterogenous Prototype & Cosine similarity \\
\cline{2-4}
& Cai \textit{et al.} \cite{icip2013} (2013) & \multicolumn{2}{c|} {Coupled least squares regression method with a local consistency constraint} \\
\cline{2-4}
& Tsai \textit{et al.} \cite{dica} (2014) & Domain adaptation based proposed DiCA & Subject-specific SVM  \\
\cline{2-4}
& Lin \textit{et al.} \cite{pami16} (2016) & Affine transformations & CNNs over Mahalanobis and Cosine scores \\
\hline
\multirow{5}{*}{Composite} & Chugh \textit{et al.} \cite{chughBTAS} (2013) & Histogram of image moments and HoG & Chi-squared distance\\
\cline{2-4}
& Han \textit{et al.} \cite{tifs2013} (2013) & MLBP of ASM features & Similarity on normalized histogram intersection \\
\cline{2-4}
& Mittal \textit{et al.} \cite{icb_paritosh} (2015) & Deep Boltzmann Machines & Neural Networks \\
\cline{2-4}
& Mittal \textit{et al.} \cite{gginfo} (2017) & HoG + DAISY & Chi-squared distance + Attribute feedback \\
\hline
\multirow{2}{*}{Both}& Klum \textit{et al.} \cite{prip} (2014) & \multicolumn{2}{c|} {SketchID- automated system based on holistic \cite{jainHFR} and component \cite{tifs2013} based algorithms} \\
\cline{2-4}
& Ouyang \textit{et al.}\cite{cvpr2016} (2016) & \multicolumn{2}{c|}{Learned a mapping to reverse the forgetting process of the eyewitness} \\
\hline
\end{tabular}
\end{center}
\vspace{-15pt}
\caption{A brief literature review of sketch-to-photo matching problem.}
\label{litRev}
\vspace{-15pt}
\end{table*}

Face recognition systems have been evolving over the past few decades, particularly with the availability of large scale databases and access to sophisticated hardware. Large scale face recognition challenges such as MegaFace \cite{megaface} and Janus \cite{janus} further provide opportunities for bridging the gap between unconstrained and constrained face recognition. However, the availability of new devices and applications continuously open new challenges. One such challenging application is matching sketches with digital face photos. In criminal investigations, eyewitnesses provide a first hand account of the event, along with a description of the appearance of the suspect based on their memory. A sketch artist interviews the eyewitness of a particular case and a sketch image of the suspect is created. Such a sketch drawn by an artist is termed as a \textit{hand-drawn sketch}. To eliminate the inter-artist variations and automate the process of sketch generation, law enforcement agencies have started using software generated \textit{composite sketches}. In such cases, the eyewitness is interviewed by an officer and a sketch is created using the drag-and-drop features available in sketch generation tools such as FACES \cite{faces}, evoFIT \cite{evofit} and IdentiKit \cite{identikit}. As shown in Figure \ref{intro1}(a), the information content in the two domains/modalities (sketch and digital image) vary significantly. The digital image is an information-rich representation whereas, the sketch image contains only the edge information and lacks texture details. Figure \ref{intro1}(b) shows real world examples of forensic hand-drawn sketch and corresponding photo pairs. Along with domain differences, variations caused by eyewitness description makes this problem further challenging. 

%This has led to the development of innovative and novel techniques for addressing real world problems which seemed far fetched few years ago. 
%------------------------------------
%Sketches are much easier to match as compared to textual information due to insufficiency of describing an image solely using words and also possible lack of image tags for the images present in the gallery. In recent times, sketch based image retrieval has garnered a lot of attention \cite{shoes, sketchnet}; however, it has been somewhat restricted to the domain of object recognition. 

%Nonetheless, it is an important aspect in the domain of faces as well, and has been used for decades for the task of face sketch recognition. 

%Automating the retrieval of top matches can result in improved results and faster processing at law enforcement agencies, therefore proving to be an important facet in computer vision. 

Traditionally, a sketch image is matched with digital mugshot images for identifying the suspect. The literature is spread across hand-drawn and composite sketch to digital photo matching \cite{Nagpal2016}, with algorithms being evaluated \cite{Jain, pami16} on viewed sketches \cite{prip,cufs, cufsf}. Viewed sketches are drawn while looking at the digital photos. Such sketches do not reflect real scenario and fail to capture the challenging nature of the problem. Choi \textit{et al.} \cite{cvprWorkshop} have established the limitations of viewed sketches and emphasized the need for new databases and algorithms imitating real scenarios.

Table \ref{litRev} summarizes the literature of facial sketch recognition which shows that both handcrafted and learned representation have been explored. Sketch recognition has traditionally been viewed as a domain adaptation task due to the cross-domain data. Such techniques can be applied for viewed sketch recognition, where the variations across different types of images is primarily governed by the changes in the domain.  However, in case of forensic sketch matching for face images, there are several factors apart from the difference in domain which make the problem further challenging, such as memory gap \cite{cvpr2016} and the bias observed due to the eye-witness \cite{icb_paritosh}. In this work, we propose a novel transform learning based formulation, \textit{DeepTransformer}, which learns meaningful coupled representations for sketch and digital images. Further, two important and challenging application scenarios are used for performance evaluation: (i) age separated digital to sketch matching (both composite and hand-drawn) and (ii) sketch to sketch matching. 
The effectiveness of the proposed formulation is evaluated on hand-drawn and forensic sketch databases, including a novel sketch database. The key contributions are:
\begin{itemize}[leftmargin=*]
\vspace{-7pt}
\item This is the first work incorporating the concept of Deep Learning in Transform Learning framework. Specifically, novel deep coupled transform learning formulations, \textit{Semi-Coupled} and \textit{Symmetrically-Coupled Deep Transform Learning}, have been presented which imbibe qualities of deep learning with domain adaption.  \vspace{-7pt}
\item This is the first work which presents sketch to sketch matching as an important, yet unattended application for law enforcement. As shown in Figure \ref{intro1}, composite and hand-drawn sketches have significant difference in their information content. Such matching can be useful for crime linking, where different methods may have been used to generate the sketches. \vspace{-7pt} %Sketch to sketch matching results are shown on 123 composite and hand-drawn sketches from ePRIP \cite{eprip} and IIIT-D Sketch \cite{iiitdSketch} databases.
\item IIIT-D CSA dataset  \footnote{Dataset will be available at www.iab-rubric.org/resources/csa.html} contains age-separated images of an individual against a sketch image, for 150 subjects. The dataset also contains 3529 digital images.
\end{itemize}

\section{Preliminaries}
Dictionary Learning has been used in literature to learn filters and feature representations \cite{lee99, Olshausen97}. For a given input $\mathbf{X}$, a dictionary $\mathbf{D}$ is learned along with the coefficients $\mathbf{Z}$:
\begin{equation} \label{KSVD}
{
\argmin_{\textit{$\mathbf{D, Z}$}} \left \|\mathbf{X} - \mathbf{DZ} \right \|_{F}^{2},\ such\ that \ \left \|\mathbf{Z}\right \|_{0} \leq \tau
}
\end{equation}
where, the $l_o$-norm imposes a constraint of sparsity on the learned coefficients. It can be observed that dictionary learning is a synthesis formulation; i.e., the learned coefficients and dictionary are able to $synthesize$ the given input $\mathbf{X}$. Ravishankar and Bresler \cite{tsp13} proposed it's analysis equivalent, termed as transform learning. It analyzes the data by learning a transform or basis to produce coefficients. Mathematically, for input data $\mathbf{X}$, it can be expressed as:
\begin{equation} \label{transform}
{
\argmin_{\textit{$\mathbf{T, Z}$}} \left \|\mathbf{TX} - \mathbf{Z} \right \|_{F}^{2},\ such\ that \left \|\mathbf{Z}\right \|_{0} \leq \tau
}
\end{equation}
where, $\mathbf{T}$ and $\mathbf{Z}$ are the transform and coefficients, respectively. Relating transform learning to the dictionary learning formulation in Equation \ref{KSVD}, it can be seen that dictionary learning is an inverse problem while transform learning is a forward problem. In order to avoid the degenerate solutions of Equation \ref{transform}, the  following formulation is proposed \cite{tsp13}: 
\begin{equation} \label{transformNew}
\begin{gathered}
\argmin_{\textit{$\mathbf{T, Z}$}} \left \|\mathbf{TX} - \mathbf{Z} \right \|_{F}^{2} +  \lambda\  \big( \epsilon \left \|\mathbf{T} \right \|_{F}^{2}  - log \det \mathbf{T} \big) 
s.t. \left \|\mathbf{Z}\right \|_{0} \leq \tau
\end{gathered}
\end{equation}
The factor `$log \det \mathbf{T}$' refers to the log-determinant regularizer \cite{logdet}, which imposes a full rank on the learned transform to prevent degenerate solutions. The additional penalty term $\left \|\mathbf{T} \right \|_{F}^{2}$ is to balance scale. In literature, an alternating minimization approach has been presented \cite{RavishankarSIAM, Ravishankar15_2} to solve the above transform learning problem, i.e.:
\begin{equation} \label{solveZ}
{
\mathbf{Z} \leftarrow \argmin_{\textit{$\mathbf{Z}$}} \left \|\mathbf{TX} - \mathbf{Z} \right \|_{F}^{2},\ such\ that \  \left \|\mathbf{Z}\right \|_{0} \leq \tau
}
\end{equation}
\begin{equation} \label{solveT}
{\mathbf{T} \leftarrow \argmin_{\textit{$\mathbf{T}$}} \left \|\mathbf{TX} - \mathbf{Z} \right \|_{F}^{2} +  \lambda\  \big( \epsilon \left \|\mathbf{T} \right \|_{F}^{2}  - log \det \mathbf{T} \big) 
}
\end{equation}
The coefficients in Equation \ref{solveZ} are updated using Orthogonal Matching Pursuit (OMP) \cite{Pati93}, and transform matrix $\mathbf{T}$ is updated using a closed form solution presented in \cite{Ravishankar15}.
%\begin{equation} \label{transform2_1}
%\mathbf{
%XX^{T}\ +\ \lambda \epsilon I\ =\ LL^{T} 
%}
%\end{equation}
%Since it is symmetric positive definite, the first step computes the Cholesky decomposition. This is followed by computing the full SVD and the update step:
%\begin{equation} \label{transform2_2}
%\mathbf{
%L^{-1}XZ^{T}\ =\ USV^{T} 
%}
%\end{equation}
%\begin{equation} \label{transform2_3}
%\mathbf{
%T\ =\ 0.5V\ \big(S\ +\ (S^{2}\ +\ 2 \lambda I )^{\frac{1}{2}} \big)\ U^{T} L^{-1}
%}
%\end{equation}
%One must notice that $\mathbf{L^{-1}}$  is easy to compute since it is a lower triangular matrix. 
The proof for convergence of the update algorithm can be found in \cite{Ravishankar15_2}. 
There is a computational advantage of transform learning over dictionary learning. The latter is a synthesis formulation, and during the test stage, for a given $x_{test}$ it needs to solve a problem of the form:
\begin{equation} \label{train}
\argmin_{\textit{${z_{test}}$}} \left \|x_{test} - \mathbf{D}z_{test} \right \|_{F}^{2},\ such\ that \  \left \|z_{test}\right \|_{0} \leq \tau
\end{equation}
This is an iterative optimization problem, and thus time consuming, whereas, transform learning is an analysis framework,  and at testing time, only the given equation is solved:
\begin{equation} \label{test}
\argmin_{\textit{$z_{test}$}} \left \|\mathbf{T}x_{test} - z_{test} \right \|_{F}^{2},\ such\ that \  \left \|z_{test}\right \|_{0} \leq \tau
\end{equation}
This can be solved using one step of hard thresholding \cite{blu}, making test feature generation very fast and real time.

\section{DeepTransformer: Proposed Coupled Deep Transform Learning}
Transform Learning has been used for several applications such as blind compressive sensing, online learning, along with image and video de-noising \cite{denoising, compressive, Ravishankar15}. This research addresses the challenging task of sketch recognition by proposing two novel formulations: semi-coupled, and symmetrically-coupled transform learning. This is the first work which incorporates a mapping function in the transform learning framework in order to reduce between-domain variations. Further, both the models have been extended to propose Semi-Coupled DeepTransformer and Symmetrically-Coupled DeepTransformer. 
%motivated by the hierarchical representation learning property of Deep Learning architectures,
\subsection{Semi-Coupled Deep Transform Learning}
As a result of varying information content of images belonging to different domains, there is a need to reduce the domain gap while performing recognition. This is often achieved by mapping the information content of one domain's data onto the other. In real world scenarios of photo to sketch matching, generally a probe sketch image is matched with a gallery of mugshot digital images. This presents the requirement of transforming data from one domain (sketch) onto the other (digital image). For such instances, where the data from only one domain is required to be mapped to the other, Semi-Coupled Transform Learning is proposed. Let $\mathbf{X_1}$ be the data of first domain and $\mathbf{X_2}$ be the data of second domain. The proposed model learns two transform matrices, $\mathbf{T_1}$ and $\mathbf{T_2}$ (one for each domain) and their corresponding features $\mathbf{Z_1}$ and $\mathbf{Z_2}$, such that the features from the first domain can be linearly mapped ($\mathbf{M}$) onto the other. Mathematically this is expressed as:
\begin{equation} \label{semi}
\begin{gathered}
\argmin_{\textit{$\mathbf{T_1, T_2, Z_1, Z_2, M}$}} \left \|\mathbf{T_{1}X_{1}} - \mathbf{Z_{1}} \right \|_{F}^{2} +\ \left \|\mathbf{T_{2}X_{2}} - \mathbf{Z_{2}} \right \|_{F}^{2} \\
+\  \lambda \big( \epsilon \left \|\mathbf{T_1}\right \|_{F}^{2} +\ \epsilon \left \|\mathbf{T_2}\right \|_{F}^{2} -\ log \det \mathbf{T_1} -\ log \det \mathbf{T_2} \big)
\\+\  \mu\left \|\mathbf{Z_{2}} - \mathbf{MZ_{1}} \right \|_{F}^{2}
\end{gathered}
\end{equation}
Equation \ref{semi} is solved using alternating minimization approach. Specifically, this equation can be decomposed into five sub-problems, one for each variable, and then each is solved individually, as explained below. \\ 
\textbf{Sub-Problem 1:}
\begin{equation} \label{semi_p1}
\argmin_{\textit{$\mathbf{T_1}$}} \left \|\mathbf{T_{1}X_{1}} - \mathbf{Z_{1}} \right \|_{F}^{2} +\ \lambda ( \epsilon \left \|\mathbf{T_1}\right \|_{F}^{2} -\ log \det \mathbf{T_1})
\end{equation}
\noindent \textbf{Sub-Problem 2:}
\begin{equation} \label{semi_p2}
\argmin_{\textit{$\mathbf{T_2}$}} \left \|\mathbf{T_{2}X_{2}} - \mathbf{Z_{2}} \right \|_{F}^{2} +\ \lambda ( \epsilon \left \|\mathbf{T_2}\right \|_{F}^{2} -\ log \det \mathbf{T_2})
\end{equation}
The solution for Equations \ref{semi_p1}, \ref{semi_p2} is similar to the one for Equation \ref{solveT}. \\ %, and can be updated by differentiating each term independently. \\
\noindent \textbf{Sub-Problem 3:}
\begin{equation} \label{semi_p3}
\begin{gathered}
\argmin_{\textit{$\mathbf{Z_1}$}} \left \|\mathbf{T_{1}X_{1}} - \mathbf{Z_{1}} \right \|_{F}^{2} +\ \mu\left \|\mathbf{Z_{2}} - \mathbf{MZ_{1}} \right \|_{F}^{2}
\\ \equiv\ \argmin_{\textit{$\mathbf{Z_1}$}} \left \|\binom{ \mathbf{T_{1}X{1}} } { \sqrt\mu \mathbf{Z_{2}} } - \binom{ \mathbf{I} } { \sqrt\mu \mathbf{M} } \mathbf{Z_{1}} \right \|_{F}^{2}
\end{gathered}
\end{equation}
\noindent \textbf{Sub-Problem 4:}
\begin{equation} \label{semi_p4}
\begin{gathered}
\argmin_{\textit{$\mathbf{Z_2}$}} \left \|\mathbf{T_{2}X_{2}} - \mathbf{Z_{2}} \right \|_{F}^{2} +\ \mu\left \|\mathbf{Z_{2}} - \mathbf{MZ_{1}} \right \|_{F}^{2}
\\ \equiv\ \argmin_{\textit{$\mathbf{Z_2}$}} \left \|\binom{ \mathbf{T_{2}X{2}} } { \sqrt\mu \mathbf{MZ_{1}} } - \binom{ \mathbf{I} } { \sqrt\mu \ \mathbf{I} } \mathbf{Z_{2}} \right \|_{F}^{2}
\end{gathered}
\end{equation}
The above two equations are least square problems with a closed form solution, and thus can be minimized for feature representations $\mathbf{Z_1}$ and $\mathbf{Z_2}$. \\
\noindent \textbf{Sub-Problem 5:}
\begin{equation} \label{semi_p5}
\argmin_{\textit{$\mathbf{M}$}} \left \|\mathbf{Z_{2}} - \mathbf{MZ_{1}} \right \|_{F}^{2}
\end{equation}
Finally, a mapping $\mathbf{M}$ is learned between the representations $\mathbf{Z_1}$ and $\mathbf{Z_2}$ by solving the above least square equation.
%\vspace{-5pt}

Inspired by the success of deep learning \cite{deepRes, imagenet, deepCNN} to model high level abstractions and learn large variations in data, this research introduces \textit{deep} transform learning. For a $k$-layered architecture, Semi-Coupled DeepTransformer can be expressed as:
\begin{equation}
\begin{gathered}
\label{semiDeep}
\argmin_\theta \Big[\sum_{j=1}^k \mathbf{\Big( \|T_1^jI_1^j - Z_1^j \|_F^2 + \|T_2^jI_2^j - Z_2^j \|_F^2} + \\ 
+\ \lambda \big( \epsilon \|\mathbf{T_1^j} \|_{F}^{2} + \epsilon \|\mathbf{T_2^j} \|_{F}^{2} -\ log \det \mathbf{T_1^j} -\ log \det \mathbf{T_2^j} \big) \Big) + \\
\mathbf{\|Z_2^k - MZ_1^k \|_F^2}\Big]
\end{gathered}
\end{equation}
where, $\mathbf{\theta = \{\mathbf{\forall_{j = 1}^ k(T_1^j, T_2^j, Z_1^j, Z_2^j), M}}\}$. $(\mathbf{T_1^j}$, $\mathbf{I_1^j}$, and $\mathbf{Z_1^j})$ and $(\mathbf{T_2^j}$, $\mathbf{I_2^j}$, and $\mathbf{Z_2^j})$ refer to the transform matrix, input, and learned representations of the $j^{th}$ layer for the two domains respectively.  
%For the first domain, $\mathbf{T_1^j}$, $\mathbf{I_1^j}$, and $\mathbf{Z_1^j}$ refer to the transform matrix, input, and learned representations for the $j^{th}$ layer. Similarly, for the second domain, $\mathbf{T_2^j}$, $\mathbf{I_2^j}$, and $\mathbf{Z_2^j}$ correspond to the transform matrix, input, and learned features for the $j^{th}$ layer. 
$\mathbf{M}$ refers to the learned linear mapping between the final representations of the $k^{th}$ layer ($\mathbf{Z_1^k, Z_2^k}$). The input to the model, $\mathbf{I_1^1}$ and $\mathbf{I_2^1}$ are $\mathbf{X_1}$ and $\mathbf{X_2}$, i.e. training data of the first and second domains, respectively. For subsequent layers, $\mathbf{I_1^j}$ and $\mathbf{I_2^j}$ correspond to the feature representations learned at the previous layers, i.e. $\mathbf{Z_1^{j-1}}$ and $\mathbf{Z_2^{j-1}}$ respectively. 
As we go deeper and increase the value of $k$, Equation \ref{semiDeep} can be solved similar to Equation \ref{semi}. The problem can be divided into ($4k$)+1 sub-problems via alternating minimization approach: separate sub-problems for solving the transform matrices ($2k$), and the learned representations ($2k$), and one for the final mapping $\mathbf{M}$. However, solving ($4k$)+1 sub-problems can be computationally expensive as the number of parameters is large. As a cost effective alternative, the proposed model can be learned with greedy layer-wise optimization.  Here we explain the layer-wise optimization for a 2-layered semi-coupled deep transform learning model (similar greedy layer-wise optimization can be followed for $k>2$).\\
%motivated by the use of greedy layer by layer training of traditional deep learning architectures, 
\noindent \textbf{Layer One:} Learn the first layer transform matrices ($\mathbf{T_1^1, T_2^1}$) for both domains, along with the representations of the input data ($\mathbf{Z_1^1, Z_2^1}$):
\begin{subequations}
\begin{equation}
\argmin_{\textit{$\mathbf{T_1^1, Z_1^1}$}} \left \|\mathbf{T_{1}^1X_{1}} - \mathbf{Z_{1}^1} \right \|_{F}^{2} 
+\ \lambda \big( \epsilon \left \|\mathbf{T_1^1}\right \|_{F}^{2} - log \det \mathbf{T_1^1} \big)
\end{equation}     
\begin{equation}
\argmin_{\textit{$\mathbf{T_2^1, Z_2^1}$}} \left \|\mathbf{T_{2}^1X_{2}} - \mathbf{Z_{2}^1} \right \|_{F}^{2} 
+\ \lambda \big( \epsilon \left \|\mathbf{T_2^1}\right \|_{F}^{2} - log \det \mathbf{T_2^1} \big)  
\end{equation}
\label{deepSemi_1}
\end{subequations}
\noindent\textbf{Layer Two:} Using the representations learned in the first layer as input, semi-coupled transform learning is applied at the second layer to obtain the transform matrices for the second layer, for both domains ($\mathbf{T_1^2, T_2^2}$):
\begin{equation} \label{deepSemi_2}
\begin{gathered}
\argmin_{\textit{$\mathbf{T_1^2, T_2^2, Z_1^2, Z_2^2, M}$}} \left \|\mathbf{T_{1}^2Z_{1}^1} - \mathbf{Z_{1}^2} \right \|_{F}^{2} +\ \left \|\mathbf{T_{2}^2Z_{2}^1} - \mathbf{Z_{2}^2} \right \|_{F}^{2} \\
+\ \lambda \big( \epsilon \left \|\mathbf{T_1^2}\right \|_{F}^{2} +\ \epsilon \left \|\mathbf{T_2^2}\right \|_{F}^{2} -\ log \det \mathbf{T_1^2} -\ log \det \mathbf{T_2^2} \big)
\\+\ \mu\left \|\mathbf{Z_{2}^2} - \mathbf{MZ_{1}^2} \right \|_{F}^{2}
\end{gathered}
\end{equation}
%At the time of testing, the feature in the first domain is generated simply by multiplying the corresponding transform, i.e. $\mathbf{z_{1,test}} = \mathbf{T_{1}x_{1,test}}$. Since, the matching is supposed to be in the second domain, we need to map the feature by: $\mathbf{z_{2,test}} = \mathbf{MT_{1}x_{1,test}}$. The thus generated feature for the second domain is compared with the training data for label assignment. 
\vspace{-15pt}
\subsection{Symmetrically-Coupled Deep Transform Learning}
In real world scenarios, a given sketch image may be matched with a dataset of different type of sketches for crime-linking. In such cases, learning a single mapping function using semi-coupled transform learning may not be useful. For such cases, symmetrically-coupled transform learning is proposed, where two linear maps are learned; one from the first domain to the second one, and the other from the second domain to the first one. This leads to the following formulation:
\begin{equation} \label{symmetric}
\begin{gathered}
\argmin_{\textit{$\mathbf{T_1, T_2, Z_1, Z_2, M_1,M_2}$}} \left \|\mathbf{T_{1}X_{1}} - \mathbf{Z_{1}} \right \|_{F}^{2} +\ \left \|\mathbf{T_{2}X_{2}} - \mathbf{Z_{2}} \right \|_{F}^{2} \\
+\ \lambda \big( \epsilon \left \|\mathbf{T_1}\right \|_{F}^{2} +\ \epsilon \left \|\mathbf{T_2}\right \|_{F}^{2} -\ log \det \mathbf{T_1} -\ log \det \mathbf{T_2} \big)
\\+\ \mu \big(\left \|\mathbf{Z_{2}} - \mathbf{M_{1}Z_{1}} \right \|_{F}^{2} +\ \left \|\mathbf{Z_{1}} - \mathbf{M_{2}Z_{2}} \right \|_{F}^{2} \big)
\end{gathered}
\end{equation}
where, $\mathbf{M_2}$ and $\mathbf{M_1}$ correspond to the mapping matrices to transform feature representations of domain two into those of domain one, and vice versa, respectively. As before, with alternating minimization, Equation \ref{symmetric} can be optimized with the help of the following sub-problems: \\
\textbf{Sub-Problem 1:}
\begin{equation} \label{symmetric_p1}
\argmin_{\textit{$\mathbf{T_1}$}} \left \|\mathbf{T_{1}X_{1}} - \mathbf{Z_{1}} \right \|_{F}^{2} +\ \lambda ( \epsilon \left \|\mathbf{T_1}\right \|_{F}^{2} -\ log \det \mathbf{T_1})
\end{equation}
\noindent \textbf{Sub-Problem 2:}
\begin{equation} \label{symmetric_p2}
\argmin_{\textit{$\mathbf{T_2}$}} \left \|\mathbf{T_{2}X_{2}} - \mathbf{Z_{2}} \right \|_{F}^{2} +\ \lambda ( \epsilon \left \|\mathbf{T_2}\right \|_{F}^{2} -\ log \det \mathbf{T_2})
\end{equation}
Updates for the transform matrices ($\mathbf{T_1, T_2}$) remain the same as shown in Equations \ref{semi_p1} and \ref{semi_p2}. \\ %, and thus can be solved in a similar manner.\\
\noindent \textbf{Sub-Problem 3:}
\begin{equation} \label{symmetric_p3}
\begin{gathered}
\argmin_{\textit{$\mathbf{Z_1}$}} \left \|\mathbf{T_{1}X_{1}} - \mathbf{Z_{1}} \right \|_{F}^{2} +\ \mu \big( \left \|\mathbf{Z_{2}} - \mathbf{M_{1}Z_{1}} \right \|_{F}^{2}
\\+\ \left \|\mathbf{Z_{1}} - \mathbf{M_{2}Z_{2}} \right \|_{F}^{2} \big)
\end{gathered}
\end{equation}
\noindent \textbf{Sub-Problem 4:}
\begin{equation} \label{symmetric_p4}
\begin{gathered}
\argmin_{\textit{$\mathbf{Z_2}$}} \left \|\mathbf{T_{2}X_{2}} - \mathbf{Z_{2}} \right \|_{F}^{2} +\ \mu \big( \left \|\mathbf{Z_{2}} - \mathbf{M_{1}Z_{1}} \right \|_{F}^{2}
\\+\ \left \|\mathbf{Z_{1}} - \mathbf{M_{2}Z_{2}} \right \|_{F}^{2} \big)
\end{gathered}
\end{equation}
The above two equations for learning the representations ($\mathbf{Z_1, Z_2}$) of the two domains are least square minimizations, and thus have closed form solutions. \\
\noindent \textbf{Sub-Problem 5:}
\begin{equation} \label{symmetric_p5}
\argmin_{\textit{$\mathbf{M_{1}}$}} \left \|\mathbf{Z_{2}} - \mathbf{M_{1}Z_{1}} \right \|_{F}^{2}
\end{equation}
\noindent \textbf{Sub-Problem 6:}
\begin{equation} \label{symmetric_p6}
\argmin_{\textit{$\mathbf{M_{2}}$}} \left \|\mathbf{Z_{1}} - \mathbf{M_{2}Z_{2}} \right \|_{F}^{2}
\end{equation}  
Similar to Equation \ref{semi_p5}, mappings ($\mathbf{M_1, M_2}$) can be learned by solving the above using least square minimization. As discussed, the DeepTransfomer is solved using alternating minimization approach. Each of the subproblems of Equations \ref{semi} and \ref{symmetric} are solved with guaranteed convergence \cite{tsp13}. Specifically, learning Z has analytical solution and transform updates are done by conjugate gradients which can only decrease. Overall, the model has monotonically decreasing cost function and therefore, will converge. 

We extend Equation \ref{symmetric} and propose symmetrically-coupled \textit{deep} transform learning where, $\mathbf{\theta = \{\mathbf{\forall_{j = 1}^ k (T_1^j, T_2^j, Z_1^j, Z_2^j), M_{1}, M_{2}}}\}$, and $\mathbf{M_{2}}$ and $\mathbf{M_{1}}$ correspond to the mapping matrices to transform feature representations of domain two into those of domain one, and vice versa, respectively. It is mathematically expressed as:
\begin{equation}
\begin{gathered}
\argmin_\theta \Big[\sum_{j=1}^k \mathbf{\Big( \|T_1^jI_1^j - Z_1^j \|_F^2 + \|T_2^jI_2^j - Z_2^j \|_F^2} + \\ 
+\ \lambda \big( \epsilon \|\mathbf{T_1^j} \|_{F}^{2} + \epsilon \|\mathbf{T_2^j} \|_{F}^{2} -\ log \det \mathbf{T_1^j} -\ log \det \mathbf{T_2^j} \big) \Big) + \\
\mathbf{\|Z_2^k - M_{1}Z_1^k \|_F^2} + \mathbf{\|Z_1^k - M_{2}Z_2^k \|_F^2}\Big] \label{symDeep}
\end{gathered}
\end{equation}
This formulation can be solved using alternating minimization approach with ($4k$+2) sub-problems where the last two sub-problems are related to learning mappings $\mathbf{M_{1}}$ and $\mathbf{M_{2}}$. However, like Semi-Coupled DeepTransformer, we optimize symmetrically coupled deep transform algorithm in a greedy layer wise manner. The optimization for a 2-layer Symmetrically-Coupled DeepTransformer is as follows:

%We extend Equation \ref{symmetric} and propose symmetrically-coupled \textit{deep} transform learning, mathematically expressed as:
%\begin{equation}
%\begin{gathered}
%\argmin_\theta \Big[\sum_{j=1}^k \mathbf{\Big( \|T_1^jI_1^j - Z_1^j \|_F^2 + \|T_2^jI_2^j - Z_2^j \|_F^2} + \\ 
%+\ \lambda \big( \epsilon \|\mathbf{T_1^j} \|_{F}^{2} + \epsilon \|\mathbf{T_2^j} \|_{F}^{2} -\ log \det \mathbf{T_1^j} -\ log \det \mathbf{T_2^j} \big) \Big) + \\
%\mathbf{\|Z_2^k - M_{1}Z_1^k \|_F^2} + \mathbf{\|Z_1^k - M_{2}Z_2^k \|_F^2}\Big] \label{symDeep}
%\end{gathered}
%\end{equation}
%where, $\mathbf{\theta = \{\mathbf{\forall_{j = 1}^ k (T_1^j, T_2^j, Z_1^j, Z_2^j), M_{1}, M_{2}}}\}$, and $\mathbf{M_{2}}$ and $\mathbf{M_{1}}$ correspond to the mapping matrices to transform feature representations of domain two into those of domain one, and vice versa, respectively.
%This formulation can be solved using alternating minimization approach with ($4k$+2) sub-problems where the last two sub-problems are related to learning mappings $\mathbf{M_{1}}$ and $\mathbf{M_{2}}$. However, like Semi-Coupled DeepTransformer, we optimize symmetrically coupled deep transform algorithm in a greedy layer wise manner. The optimization for a 2-layer Symmetrically-Coupled DeepTransformer is as follows:

\noindent \textbf{Layer One:}
\begin{subequations}
\begin{equation}
\argmin_{\textit{$\mathbf{T_1^1, Z_1^1}$}} \left \|\mathbf{T_{1}^1X_{1}} - \mathbf{Z_{1}^1} \right \|_{F}^{2} 
+\ \lambda \big( \epsilon \left \|\mathbf{T_1^1}\right \|_{F}^{2} - log \det \mathbf{T_1^1} \big)
\end{equation}     
\begin{equation}
\argmin_{\textit{$\mathbf{T_2^1, Z_2^1}$}} \left \|\mathbf{T_{2}^1X_{2}} - \mathbf{Z_{2}^1} \right \|_{F}^{2} 
+\ \lambda \big( \epsilon \left \|\mathbf{T_2^1}\right \|_{F}^{2} - log \det \mathbf{T_2^1} \big)  
\end{equation}
\label{deepSymmetric_1}
\end{subequations}
\noindent \textbf{Layer Two:}
\begin{equation} \label{deepSymmetric2}
\begin{gathered}
\argmin_{\textit{$\mathbf{T_1^2, T_2^2, Z_1^2, Z_2^2, M}$}} \left \|\mathbf{T_{1}^2Z_{1}^1} - \mathbf{Z_{1}^2} \right \|_{F}^{2} +\ \left \|\mathbf{T_{2}^2Z_{2}^1} - \mathbf{Z_{2}^2} \right \|_{F}^{2} \\
+\ \lambda \big( \epsilon \left \|\mathbf{T_1^2}\right \|_{F}^{2} +\ \epsilon \left \|\mathbf{T_2^2}\right \|_{F}^{2} -\ log \det \mathbf{T_1^2} -\ log \det \mathbf{T_2^2} \big)
\\+\ \mu (\left \|\mathbf{Z_{2}^2} - \mathbf{M_{1}Z_{1}^2} \right \|_{F}^{2}+\ \left \|\mathbf{Z_{1}^2} - \mathbf{M_{2}Z_{2}^2} \right \|_{F}^{2})
\end{gathered}
\end{equation}
The first layer learns the low level representation of each domain independently, while the second layer learns the high level representations and mapping between the representations of the two domains/modalities. %In other words, the second layer thus builds upon the first layer features by learning higher level representations from the learned low level features. 
The proposed model thus encodes domain specific features, followed by features incorporating the between-domain variations. 

\subsection{DeepTransformer for Sketch Recognition}
The proposed two layer DeepTransformer is used for performing face sketch matching. For semi-coupled DeepTransformer, the following steps are performed: \\
\noindent \textbf{Training:} Given a set of sketch and digital training pairs, $\mathbf{X_s, X_d}$, transform matrices ($\mathbf{T_s^1, T_d^1, T_s^2, T_d^2}$) and coefficient vectors ($\mathbf{Z_s^1, Z_d^1, Z_s^2, Z_d^2}$) are learned using Equation \ref{semiDeep}, along with a mapping, $\mathbf{M}$, between $\mathbf{Z_s^2, Z_d^2}$. A two hidden layer neural network classifier is trained to make identification decisions.\\
\noindent \textbf{Testing:} For a given probe sketch image, $x_{sTest}$, the first and second layer feature representations are extracted using the learned transform spaces:
\begin{equation} 
\label{test_1}
z_{sTest}^{1} = \mathbf{T_s^1}x_{sTest}; \ z_{sTest}^{2} = \mathbf{T_s^2}z_{sTest}^{1}
\end{equation}
%\begin{equation} 
%\label{test_2}
%z_{sTest}^{2} = \mathbf{X_s^2}z_{sTest}^{1}
%\end{equation}
The mapping $\mathbf{M}$ is used to transform the feature vector onto the digital image space, i.e, $z_{dTest}^{2} = \mathbf{M_1}z_{sTest}^{2}$. The feature representation of the sketch (probe) in the digital image feature space, $z_{dTest}$ is now used for performing recognition using the trained neural network. For sketch to sketch matching (i.e. cases where mappings to and from different modalities are required), similar steps can be followed for utilizing symmetrically-coupled deep transform learning.

\begin{table*}
\begin{center}
\small
\centering
\begin{tabular}{|M{3.em}|M{10.2em}|M{5em}||M{7.6em}|M{6.5em}|M{6.15em}|M{2.8em}|M{2.8em}| } 
%\begin{tabular}{ |c|c|c||c|c|c|c|c| } 
\hline
\textbf{Protocol} & \textbf{Gallery Type} & \textbf{Probe (Sketch)} & \textbf{Databases for Feature Learning} & \textbf{Test Database} & \textbf{No. of Training Pairs} & \textbf{Size of Gallery} & \textbf{Size of Probe}  \\
\hline
\hline
\multicolumn{8}{|c|}{\textbf{Matching sketch to age-separated digital images: Semi-Coupled DeepTransformer}}  \\
\hline
\hline
P1 & Younger age images & Composite & \multirow{5}{*}{\parbox{7.3em}{\centering CUFS, CUFSF, e-PRIP, PRIP-VSGC, IIIT-D Viewed and Semi-viewed}} & CSA & 2129 & 875 & 90\\ 
\cline{1-3} 
\cline{5-8}
P2 & Same age images & Composite &  & CSA & 2129 & 90 & 90 \\ 
\cline{1-3} 
\cline{5-8}
P3 & Older age images & Composite &   & CSA & 2129 & 1044 & 90 \\
\cline{1-3} 
\cline{5-8}
P4 & \multirow{2}{*}{\parbox{10.2em}{\centering{Large-scale image dataset}}} & Composite &  & CSA & 2129 & 7165 & 90 \\
\cline{1-1}
\cline{3-3}
\cline{5-8}
P5 &  & Forensic &  & IIIT-D Forensic  & 2129 & 7265 & 190 \\
\hline
\hline
\multicolumn{8}{|c|}{\textbf{Sketch to sketch matching: Symmetrically-Coupled DeepTransformer}}  \\
\hline
\hline
P6 & Composite  & Hand-drawn &  CUFS, e-PRIP & CUFS, e-PRIP & 50 & 73 & 73\\ 
 \hline
P7 & Hand-drawn & Composite & CUFS, e-PRIP & CUFS, e-PRIP & 50 & 73 & 73 \\
\hline
\end{tabular}
\vspace{-5pt}
\caption{Details of experimental protocols. For P6-P7, unseen training and testing partitions are used from the CUFS and e-PRIP datasets (both contain sketches pertaining to AR dataset).}
\label{protocol}
\end{center}
\vspace{-20pt}
\end{table*}

\section{Databases and Experimental Protocol}
 %\noindent \textbf{Proposed Composite Sketch with Age Variations (CSA)}: 
Face sketch databases \cite{prip,cufs} generally comprise of viewed sketches, either hand-drawn or composite. Viewed sketches are created by looking at a digital image and sketching it simultaneously. This fails to capture the uncertainty in the recall process that humans encounter or the variations in characteristics, like a hairstyle modification, that are generally present between a sketch and digital image acquired at different times. In this research, we utilize a novel IIIT- D \textit{Composite Sketch with Age variations (CSA)} dataset \cite{cvprw}, which is the \textbf{first} publicly available dataset containing multiple age-separated digital images for a given sketch image. %In our previous work \cite{cvprw}, a smaller portion  presented IIIT-D Composite Face Sketch dataset, which contained two digital images for a given sketch image. %Unlike viewed sketches, the dataset contains images for different ages, resulting in images incorporating a change in setting and not being exactly same. 
Inspired by Bhatt \textit{et al.} \cite{iiitdSketch}, the human forgetting process is incorporated by creating semi-forensic composite sketches. The user is shown the digital image of a subject for a few minutes, and is asked to create the composite sketch after a period of 30 minutes based on his/her memory. The database consists of 3529 sketches and digital face images pertaining to 150 individuals. Out of the 150 subjects, 52 are selected from the FG-NET Aging Database \cite{fgnet}, 82 are selected from IIIT-D Aging Database \cite{iiitdAging}, and the remaining subjects are collected from the Internet. The composite sketch images are created using FACES \cite{faces}, a popular software to generate photo-like composite sketches. %containing 4400 individual facial features including distinguishing marks such as piercings and moles. 

In IIIT-D CSA dataset, the digital images span over an age range of 1 to 65 years. For each subject, an image is chosen from the middle of his/her age range and a corresponding sketch image is generated. Following this, each subject's digital images are divided into three categories: 

\noindent \textbf{(i) Younger age group:} This category models the scenario when the digital images (gallery) are younger than the probe sketch image. This set contains 1713 digital images. 

\noindent \textbf{(ii) Same age group:} This category represents the scenario when the age of an individual is similar in both digital image (gallery) and sketch image (probe). A total of 150 digital images exist in this set.

\noindent \textbf{(iii) Older age group:} This category imitates the scenario when the digital images (gallery) of the individuals are at an age older than the sketch. It consists of 1516 digital images.

\begin{figure}
\center
\includegraphics[width = 3.1in]{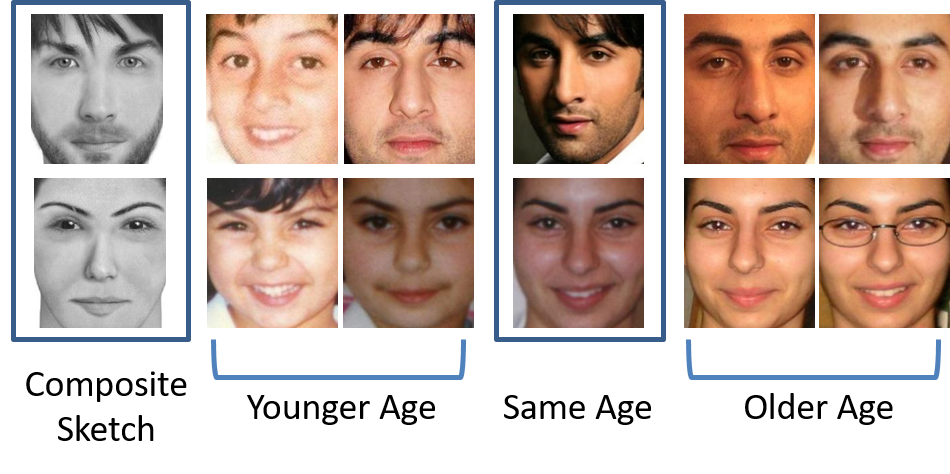}
\vspace{-10pt}
\caption{Sample images from the IIIT-D CSA database showing a sketch and age-separated face images of two subjects.}
\label{csaFig}
\vspace{-15pt}
\end{figure}

Overall, IIIT-D CSA consists of 150 composite sketch images, one for each subject, and 3379 digital images belonging to different age categories. Figure \ref{csaFig} shows sample images from the dataset. %This dataset will be made publicly available to researchers along with the train-test partitions (40\% subjects in training and remaining 60\% unseen in testing). %\newline
 
%\noindent\textbf{Existing Databases:} 

Apart from IIIT-D CSA, we have also used viewed hand-drawn sketch and digital image pairs from CUHK Face Sketch Dataset (\textbf{CUFS}) \cite{cufs} (311 pairs of students and AR dataset \cite{ar}), CUHK Face Sketch FERET Dataset (\textbf{CUFSF}) \cite{cufsf} (1194 pairs), and \textbf{IIIT-D Sketch dataset} \cite{iiitdSketch}. IIIT-D dataset contains viewed (238 pairs), semi-viewed (140 pairs), and forensic sketches (190 pairs). Composite sketches from PRIP Viewed Software-Generated Composite database (\textbf{PRIP-VSGC}) \cite{prip} and extended-PRIP Database (\textbf{e-PRIP}) \cite{eprip} (Indian user set) are also used. %Sketches generated by an Indian user (e-PRIP) and an American user (PRIP-VSGC) have been utilized in our experiments. 

\noindent \textbf{Experimental Protocol:} To evaluate the efficacy of the proposed formulations two challenging problems are considered: sketch matching against age-separated digital images (semi-coupled DeepTransformer) and sketch to sketch matching (symmetrically coupled DeepTransformer). Since this is the first research that focuses on sketch to sketch matching, as well as sketch to age-separated digital matching, we have created seven different experimental protocols to understand the performance with individual cases. These protocols are classified according to the two case studies and the details are summarized in Table \ref{protocol}. 

\noindent \textbf{1. Matching Sketch to Age-Separated Digital Images:} CSA test set and IIITD Forensic hand-drawn database have been used to evaluate the performance of the proposed model.  Inspired from real life scenarios, the test set is divided into a gallery and probe set. The gallery contains the digital images while the probe contains the sketch image. The first three protocols evaluate the effect of age difference on the recognition performance, and the next two protocols (P4 and P5) analyze the difference in performance on matching forensic and composite sketches with large scale digital image gallery. Since sketch to digital image matching experiment involves one way mapping, the results are demonstrated with Semi-Coupled DeepTransformer.

\noindent \textbf{2. Sketch to Sketch Matching:} In real world crime scene linking application, one might want to match a hand-drawn sketch against a database of composite sketches, or the other way around. Therefore, for this experiment, the proposed Symmetrically-Coupled DeepTransformer is used. CUFS dataset contains hand-drawn sketch images for the AR dataset (123 subjects), while e-PRIP contains composite sketches generated by a sketch artist for the same. The following two experiments with protocols P6 and P7 are performed: \textbf{(i)} composite to hand-drawn sketch, and \textbf{(ii)} hand-drawn to composite sketch matching.

\section{Results and Observations}
Effectiveness of DeepTransformer is evaluated with 
%Since the proposed DeepTransformer model is feature independent, it's effectiveness is evaluated with 
multiple input features, namely Dense Scale Invariant Feature Transform (DSIFT) \cite{dsift}, Dictionary Learning (DL) \cite{lee99}, Class Sparsity based Supervised Encoder (L-CSSE) \cite{majumdar16}, Light CNN \cite{lightCnn}, and VGG-Face \cite{vgg}. To analyze the effect of depth in this formulation, the results are computed with single layer (low level features) and with two layers (high level features) of DeepTransformer. Two kinds of comparative experiments are performed. The first one compares the performance of one layer and two layers deep transform learning algorithms with two classifiers, i.e., Euclidean distance and neural network. The second comparison is performed with existing algorithms like Semi-Coupled Dictionary Learning algorithm (SCDL) \cite{scdl} and Multi-Modal Sharable and Specific feature learning algorithm (MMSS) \cite{mmss}. Both the techniques have been used in literature for performing cross-domain recognition, wherein the former is a coupled dictionary learning based approach (synthesis technique), and the latter incorporated transform learning with convolutional neural networks for addressing cross-domain recognition. Comparison has also been drawn with state-of-the-art sketch recognition algorithms, namely MCWLD \cite{iiitdSketch} and GSMFL \cite{pami16}, and a commercial-off-the-shelf system (COTS), Verilook \cite{verilook}. 
%All the existing models have been used as is for obtaining classification accuracies on the given experimental protocols. 
In all the experiments, for training the networks, data augmentation is performed on the gallery images to increase per-class samples by varying the illumination and flipping the images along the y-axis. The key observations from experimental results are:
%Data augmentation is performed on the gallery images to increase per-class samples.  

\noindent \textbf{Performance with Different Features:} Tables \ref{sketchRes2} and \ref{csaRes} present the rank-10 identification accuracies of DeepTransformer with different features, for both applications of sketch matching: sketch to sketch matching and sketch to photo matching. Table \ref{csaRes} presents the accuracies for sketch to digital image matching, where the proposed Semi-Coupled DeepTransformer has been used. The results show that DeepTransformer enhances the performance of existing feature extraction techniques by at least 10\% as compared to Euclidean distance matching, and at most 22\% when neural network (NNET) is used for classification. Upon comparing accuracies across features, it is observed that DeepTransformer achieves the best results with L-CSSE features for all protocols. Similar results can be seen from Table \ref{sketchRes2} where Symmetrically-Coupled DeepTransformer has been used for sketch to sketch matching. Experimentally, it can be observed that providing class-specific features to DeepTransformer results in greater improvement. L-CSSE is a supervised deep learning model built over an autoencoder. The model incorporates supervision by adding a $ l_{2,1}$ norm regularizer during the feature learning to facilitate class-specific feature learning. The model utilizes both global and local facial regions to compute feature vector and has been shown to achieve improved results for existing face recognition problems. Further, we also observe that L-CSSE encodes the high frequency features in both local and global regions which are pertinent to digital face to sketch matching. Moreover, improved performance is observed for hand-crafted, as well as representation learning based features, thus promoting the use of DeepTransformer for different types of feature extraction techniques and input data.  

\noindent \textbf{Comparison with Existing Approaches:} Table \ref{csaResExisting} shows that the proposed, DeepTransformer with L-CSSE features outperforms existing algorithms for both the applications of sketch recognition. In case of sketch to digital image matching, with younger age protocol ({P1}), Semi-Coupled DeepTransformer attains a rank-10 accuracy of \textbf{42.6\%}, which is at least 15\% better than existing algorithms, and around 24\% better than COTS. Similar trends are observed for P2 and P3 protocols, where the proposed DeepTransformer outperforms existing techniques and the commercial-off-the-shelf system by a margin of at least 13\% and 11\% respectively. Additionally, the matching accuracy achieved by the proposed Symmetrically-Coupled DeepTransformer exceeds existing techniques for the task of sketch to sketch matching as well (P6, P7).  An improvement of at least 14\% and at most 20\% is seen with the proposed DeepTransformer (L-CSSE as feature) for the given protocols. This accentuates the use of DeepTransformer for addressing the problem of real world sketch matching.

\begin{table}
\begin{center}
\centering
\small
\begin{tabular}{ |l|M{4em}|M{4em}|M{4em}|M{4em}| }
\hline
%\multicolumn{5}{|c|}{\textbf{Rank - 10 Identification Accuracy (\%)}} \\
%\hline
%\hline
\multirow{2}{*}{\textbf{Features}} & \textbf{Euclidean} & \multirow{2}{*}{\textbf{NNET}} & \multicolumn{2}{c|}{\textbf{DeepTransformer}} \\ % (1-Layer)} & \textbf{Deep-Transformer (2-Layer)} \\
\cline{4-5}
& \textbf{Distance} & & \textbf{1-Layer} & \textbf{2-Layer} \\
\hline
\hline
\multicolumn{5}{|c|}{\textbf{Gallery with Younger Age Digital Images (P1)}} \\
\hline
\hline
DSIFT & 8.9 & 15.6 & 26.7 & 27.8 \\
\hline
%HOG & & & & \\
% \hline
%LBP & 5.6 & 16.7 & 17.8 & 18.9 \\
% \hline
DL & 2.2 & 14.4 & 17.8 & 17.8 \\
\hline
VGG & 1.1 & 11.1 & 12.2 & 12.2 \\
\hline
Light CNN & 8.9 & 12.2 & 30.0 & 27.8 \\
\hline
L-CSSE & 14.4 & 19.7 & 34.1 & \textbf{42.6} \\
%\hline
%MCWLD & \multicolumn{4}{c|}{26.8} \\
%\hline
%GSMFL & \multicolumn{4}{c|}{25.2} \\
%\hline
%Verilook & \multicolumn{4}{c|}{17.78} \\
\hline
\hline
\multicolumn{5}{|c|}{\textbf{Gallery with Same Age Digital Images (P2) }} \\
\hline
\hline
DSIFT & 7.8 & 26.7 & 25.6 & 27.8 \\
\hline
%HOG & & & & \\
%\hline
%LBP & 5.6 & 16.7 & 17.8 & 17.8 \\
% \hline
DL & 1.1 & 13.3 & 15.6 & 17.8 \\
\hline
VGG & 2.2 & 12.2 & 14.4 & 14.4 \\
\hline
Light CNN & 11.1 & 25.6 & 32.2 & 34.4 \\
\hline
L-CSSE & 16.3 & 30.2 & 37.7 & \textbf{44.2} \\
%\hline
%MCWLD & \multicolumn{4}{c|}{30.7} \\
%\hline
%GSMFL & \multicolumn{4}{c|}{29.3} \\
%\hline
%Verilook & \multicolumn{4}{c|}{15.56} \\
\hline
\hline
\multicolumn{5}{|c|}{\textbf{Gallery with Older Age Digital Images (P3)}} \\
\hline
\hline
DSIFT & 5.6 & 21.1 & 23.3 & 24.4 \\
\hline
%HOG & & & & \\
% \hline
%LBP & 3.3 & 12.2 & 16.7 & 18.9 \\
% \hline
DL & 2.2 & 13.3 & 17.8 & 18.9 \\
\hline
VGG & 2.2 & 11.1 & 12.2 & 12.2 \\
\hline
Light CNN & 7.8 & 20.0 & 24.4 & 28.9 \\
\hline
L-CSSE & 9.9 & 20.0 & 28.9 & \textbf{36.0}\\
\hline
%MCWLD & \multicolumn{4}{c|}{23.9} \\
%\hline
%GSMFL & \multicolumn{4}{c|}{23.1} \\
%\hline
%Verilook & \multicolumn{4}{c|}{12.2} \\
%\hline
\end{tabular}
\vspace{-5pt}
\caption{Rank-10 accuracies (\%) for protocols P1 to P3 using proposed Semi-Coupled DeepTransformer.}
\label{csaRes}
\end{center}
\vspace{-25pt}
\end{table}

%Protocol 6 (gallery: indian, probe: hd) and protocol 7 (gallery: hd, probe: indian)
\begin{table*}
\begin{center}
\small
\centering
\begin{tabular}{ |l||M{4.6em}|M{4.6em}|M{4.6em}|M{4.6em}||M{4.6em}|M{4.6em}|M{4.6em}|M{4.6em}|M{4.6em}| }
\hline
& \multicolumn{4}{c||}{\textbf{Gallery: Composite, Probe: Hand-drawn (P6)}} & \multicolumn{4}{c|}{\textbf{Gallery: Hand-drawn, Probe: Composite (P7)}} \\
\hline
\hline

\multirow{2}{*}{\textbf{Features}} & \textbf{Euclidean} & \multirow{2}{*}{\textbf{NNET}} & \multicolumn{2}{c||}{\textbf{DeepTransformer}}  & \textbf{Euclidean} & \multirow{2}{*}{\textbf{NNET}} & \multicolumn{2}{c|}{\textbf{DeepTransformer}} \\ % (1-Layer)} & \textbf{Deep-Transformer (2-Layer)} \\
\cline{4-5}
\cline{8-9}
& \textbf{Distance} & & \textbf{1-Layer} & \textbf{2-Layer} & \textbf{Distance} & & \textbf{1-Layer} & \textbf{2-Layer} \\
\hline
DSIFT & 4.1 & 16.4 & 24.7 & 28.8 & 2.7 & 13.7 & 23.3 & 30.1\\
\hline
DL & 4.1 & 15.1 & 17.8 & 19.2 & 6.9 & 16.4 & 19.2 & 20.6 \\
\hline
VGG &  6.9 & 12.3 & 24.7 & 27.4 & 6.9 & 15.1 & 19.2 & 20.6 \\
\hline
Light CNN &  8.2 & 15.1 & 26.0 & 30.1 & 8.2 & 16.4 & 20.6 & 28.8 \\
\hline
L-CSSE & 10.9 & 17.8 & 28.4 & \textbf{31.5} & 10.9 & 20.9 & 31.5 & \textbf{33.6} \\
\hline
\end{tabular}
\vspace{-5pt}
\caption{Rank-10 accuracies (\%) for sketch to sketch matching (P6, P7) using Symmetrically-Coupled DeepTransformer.}
\label{sketchRes2}
\end{center}
\vspace{-20pt}
\end{table*}

\begin{table}
\begin{center}
\small
\centering
\begin{tabular}{ |l|M{2em}|M{2em}|M{2em}|M{2em}|M{2em}| }
\hline
%\multicolumn{4}{|c|}{\textbf{Rank - 10 Identification Accuracy (\%)}} \\
%\hline
%\hline
\textbf{Algorithm} & \textbf{P1} & \textbf{P2} & \textbf{P3} & \textbf{P6} & \textbf{P7} \\
\hline
\hline
MCWLD \cite{iiitdSketch} & 26.8 & 30.7 & 24.4 & 16.5 & 19.2 \\
\hline
GSMFL \cite{pami16} & 25.2 & 29.3 & 23.3 & 16.5 & 19.2 \\
\hline
SCDL \cite{scdl} & 23.3 & 25.6 & 18.9 & 15.1 & 13.7 \\
\hline
MMSS \cite{mmss} & 22.2 & 27.8 & 21.1 & 13.3 & 15.1 \\
\hline
Verilook (COTS) \cite{verilook} & 17.8 & 16.6 & 12.2 & 10.9 & 13.7 \\
\hline
DeepTransformer & \textbf{42.6} & \textbf{44.2} & \textbf{36.0} & \textbf{31.5} & \textbf{33.6}\\
(with L-CSSE)  &&&&&\\
\hline
\end{tabular}
\vspace{-5pt}
\caption{Rank-10 accuracies (\%) comparing proposed DeepTransformer with existing algorithms and COTS.}
\label{csaResExisting}
\end{center}
\vspace{-25pt}
\end{table}

%\begin{table}
%\begin{center}
%\small
%\centering
%\begin{tabular}{ |M{5.3em}|M{3.5em}|M{3.5em}|M{3.5em}|M{3.5em}| }
%\hline
%%\multicolumn{4}{|c|}{\textbf{Rank - 10 Identification Accuracy (\%)}} \\
%%\hline
%%\hline
%\textbf{Algorithm} & \textbf{P6} & \textbf{P7} & \textbf{P8} & \textbf{P9} \\
%\hline
%\hline
%MCWLD \cite{iiitdSketch} & 19.2 & 16.5 & 19.2 & 19.2 \\
%\hline
%GSMFL \cite{pami16} & 19.2 & 16.5 & 17.8 & 19.2 \\
%\hline
%SCDL \cite{scdl} & 19.2 & 15.1 & 16.4 & 13.7 \\
%\hline
%MMSS \cite{mmss} & 16.4 & 13.3 & 16.4 & 15.1 \\
%\hline
%Verilook \cite{verilook} & 15.1 & 10.9 & 17.8 & 13.7\\
%\hline
%Proposed & \textbf{34.8} & \textbf{31.5} & \textbf{35.3} & \textbf{33.6} \\
%\hline
%\end{tabular}
%\vspace{-5pt}
%\caption{Rank-10 accuracies (\%) for sketch to sketch matching. }
%\label{resSketchExisting}
%\end{center}
%\vspace{-25pt}
%\end{table}
\noindent \textbf{Effect of Layer-by-Layer Training and Number of Layers:} We compare the performance of the proposed DeepTransformer with and without layer-by-layer training (i.e. Equations \ref{semiDeep} and \ref{symDeep} for direct solving for $k=2$ and layer-by-layer training as per Equations \ref{deepSemi_1}-\ref{deepSemi_2} and \ref{deepSymmetric_1}-\ref{deepSymmetric2}). On a 108-core server with 256GB RAM, for protocols P1 to P3, training Semi-Coupled DeepTransformer with layer-by-layer training requires 142 seconds which is 12 seconds faster than without layer-by-layer training. For both the cases, for $k=2$, the rank-10 accuracies are same which shows that layer-by-layer training is cost effective. We also analyze the effect of number of layers and, as shown in Tables \ref{sketchRes2} and \ref{csaRes}, 1-8\% improvement in rank-10 accuracy is observed for different protocols upon going deeper. %from one layer to two layers. %These results motivate the use of a deep architecture on transformation learning, in order to learn improved representations.  
%all protocols (1-9)  upon using the 2-layer architecture. Improvement as high as 7\% is observed, strengthening our motivation of going deeper for learning feature abstraction. 
%For sketch to digital image matching, an improvement in performance can be observed in all the given protocols (1-4) on all features, therefore motivating the need to use a deep architecture. 
%
%Similarly, for experiments pertaining to sketch to sketch matching, i.e., protocol-6 to protocol-9, an improvement of at least 1\% and at most 6\% is observed while going deeper.

\begin{figure}[t]
\centering
\captionsetup[subfigure]{labelformat=empty}
\subfloat[]{\includegraphics[clip, trim=2cm 6.75cm 1cm 9cm, width=3.2in]{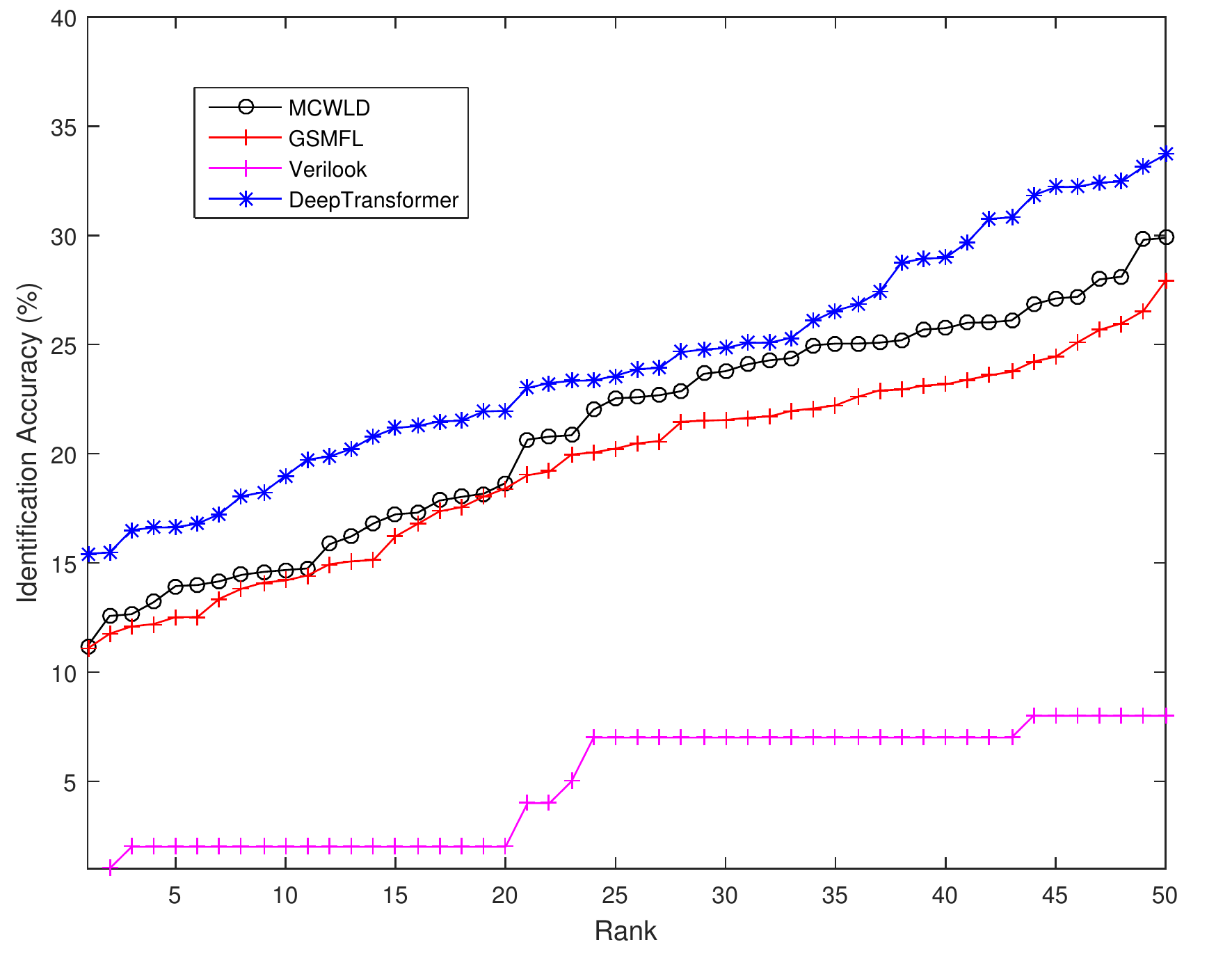}} \\
\vspace{-5pt}
\subfloat[]{\includegraphics[clip, trim=2cm 6.75cm 1cm 9cm, width=3.2in]{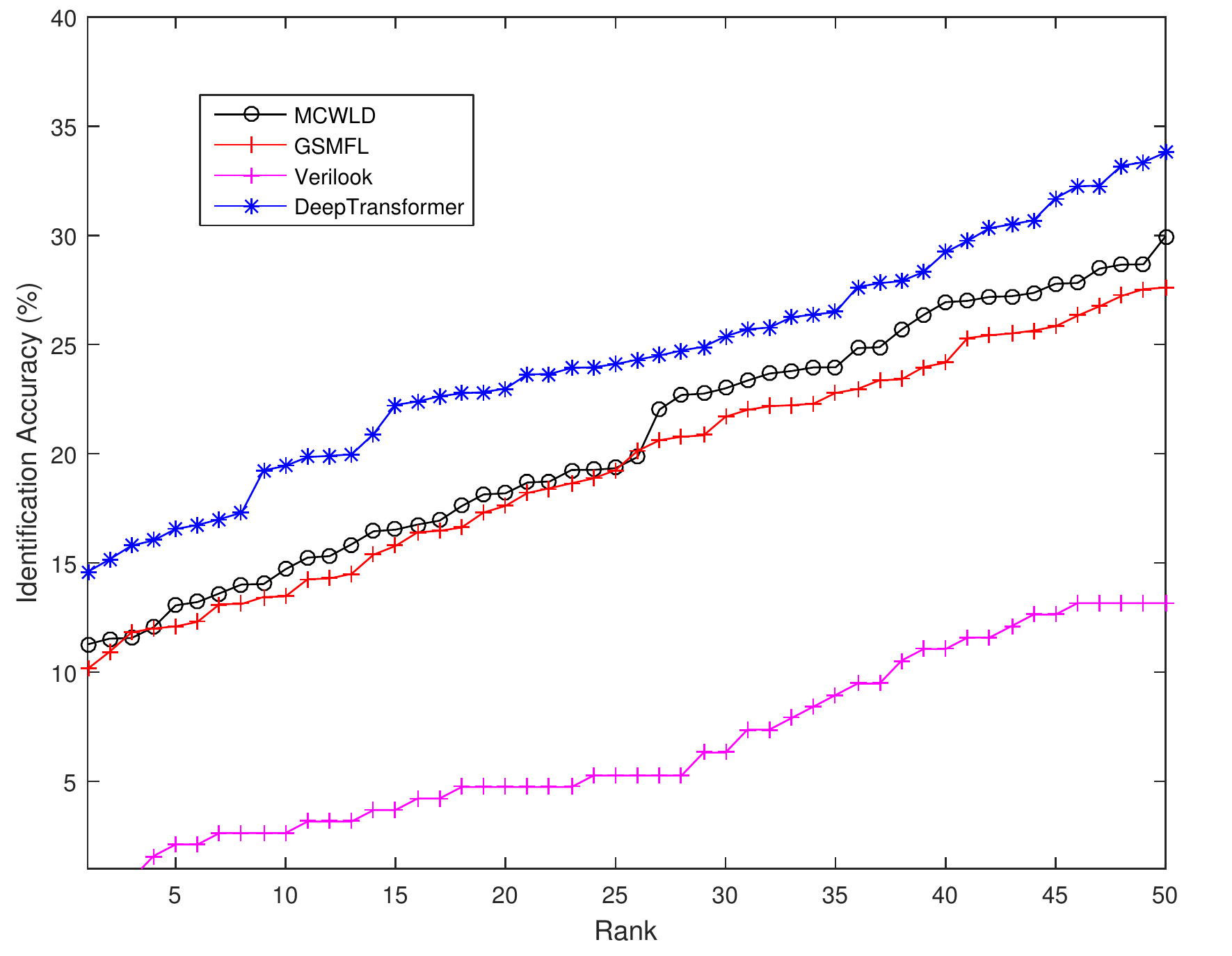}}
\vspace{-5pt}
\caption{CMC curves for P4 and P5 experiments: (a) CSA, and (b) IIIT-D Forensic datasets.}
\label{cmc}
\vspace{-5pt}
\end{figure}

\noindent \textbf{Performance on Large-Scale Dataset:} The performance of the proposed DeepTransformer has also been evaluated for a large-scale real world dataset using protocols P4 and P5. Figure \ref{cmc} presents the Cumulative Match Characteristic curves (CMCs) for IIIT-D CSA composite and IIIT-D Forensic hand-drawn sketch database respectively. The proposed Semi-Coupled DeepTransformer achieves a rank-50 accuracy of 33.7\%, which is an improvement of at least 5\% from other algorithms on IIIT-D CSA dataset. Similar results can be observed on the forensic sketches as well.

The experimental results showcase the efficacy of the proposed DeepTransformer, in terms of the improvement in identification accuracies with different features, and in comparison with other existing models. The results suggest that the DeepTransformer is robust to the type of feature and has the ability to learn over varying input spaces. Moreover, efficient training of the symmetrically coupled DeepTransformer with as few as 50 digital-sketch pairs (P6 and P7) motivate the use of the proposed architecture for small sample size problems as well. The evaluation on different real world protocols further strengthens the usage of the proposed model for addressing cross domain matching tasks. 

\section{Conclusion}
%This research focuses on the challenging problem of face sketch recognition and proposes a novel transform learning based formulation, called as \textit{DeepTransformer}. Two models: Semi-Coupled and Symmetrically-Coupled Deep Transform Learning have been presented, both of which aim to reduce the variations between two domains. The highlight of the proposed formulation is that it provides the flexibility of using any feature extractor and classifier in the framework. The proposed models have been evaluated with real world scenarios of age-separated digital image to sketch matching and sketch to sketch matching. Due to the lack of an existing database created in forensic scenarios, this work also presents the Composite Sketch with Age variations database of 150 subjects. Comparison with existing state-of-the-art algorithms and commercial-off-the-shelf system further instantiates the efficacy of DeepTransformer. 

This research focuses on the challenging problem of face sketch recognition and proposes a novel transform learning based formulation, called as \textit{DeepTransformer}. Two models: Semi-Coupled and Symmetrically-Coupled DeepTransformer have been presented, both of which aim to reduce the variations between two domains. The highlight of the proposed formulation is that it provides the flexibility of using an existing feature extractor and classifier in the framework. The proposed DeepTransfomer is evaluated with real world scenarios of age-separated digital image to sketch matching and sketch to sketch matching. Results are also shown on the IIIT-D Composite Sketch with Age variations database of 150 subjects. Comparison with existing state-of-the-art algorithms and commercial-off-the-shelf system further instantiates the efficacy of both the semi-coupled and symmetrically coupled variants of the purposed DeepTransformer. 

\section{Acknowledgment}
This research is partially supported by MEITY (Government of India), India. M. Vatsa, R. Singh, and A. Majumdar are partially supported through Infosys Center for Artificial Intelligence. S. Nagpal is partially supported through TCS PhD Fellowship. The authors acknowledge T. Chugh for his help in database creation. R. Singh also thank NVIDIA Corp. for Tesla K40 GPU for research.
 
{\small
\bibliographystyle{ieee}
\bibliography{egbib}
}

\end{document}